\documentclass[11pt]{scaleai-paper}

\usepackage{amsmath}
\usepackage{amsfonts}
\usepackage{amssymb}
\usepackage{booktabs}
\usepackage{multirow}
\usepackage{subcaption}
\usepackage{float}
\usepackage{placeins}
\usepackage[square,numbers,sort&compress]{natbib}
\usepackage{xspace}
\usepackage{url}
\usepackage[colorlinks=true,linkcolor=scaleLink,citecolor=scaleLink,urlcolor=scaleLink]{hyperref}
\usepackage[capitalise,nameinlink]{cleveref}

\let\svthefootnote\thefootnote
\newcommand\freefootnote[1]{%
  \let\thefootnote\relax%
  \footnotetext{#1}%
  \let\thefootnote\svthefootnote%
}

\graphicspath{{figures/}{figures/out/}{../figures/}{../figures/out/}}

\papertype{Scale AI Research}
\contact{\texttt{\{manasi.sharma, yue.zhang\}@scale.com, chenguangwang@ucsc.edu}}

\title{ChainWorld: Composing Long-Horizon Desktop Workloads from Atomic OSWorld Tasks}

\author[1,2*]{Vincent Siu}
\author[1]{Manasi Sharma}
\author[3]{Dawn Song}
\author[1]{Daniel Yue Zhang}
\author[1,2]{Chenguang Wang}
\author[1]{Ying Liu}

\affil[1]{Scale AI}
\affil[2]{University of California, Santa Cruz}
\affil[3]{University of California, Berkeley}

\begin{document}

\maketitle
\freefootnote{${}^*$Work done during an internship at Scale AI.}

\begin{abstract}
Computer use agents are evaluated almost exclusively on atomic desktop tasks, but realistic desktop work requires sustaining state across multiple objectives. We study this gap with ChainWorld, which composes atomic OSWorld tasks into long horizon desktop workloads through directional compatibility search while preserving the source evaluators. The resulting workload contains 347 chains of length two to four and compares two renderings of the same task sequence. In \textbf{single turn} evaluation, all tasks are presented together in one prompt. In \textbf{multi turn} evaluation, tasks are revealed one at a time. Across four current computer use agents, maximum chain completion is 31\%. Multi turn evaluation improves completion for three models, but both protocols remain challenging. The two protocols also expose different failure profiles. Single turn failures concentrate on artifact precision, while multi turn failures more often reflect session management problems such as fragmented progress and later turn disengagement. 
\end{abstract}

\section{Introduction}

\begin{figure*}[t]
\centering
\includegraphics[width=\textwidth]{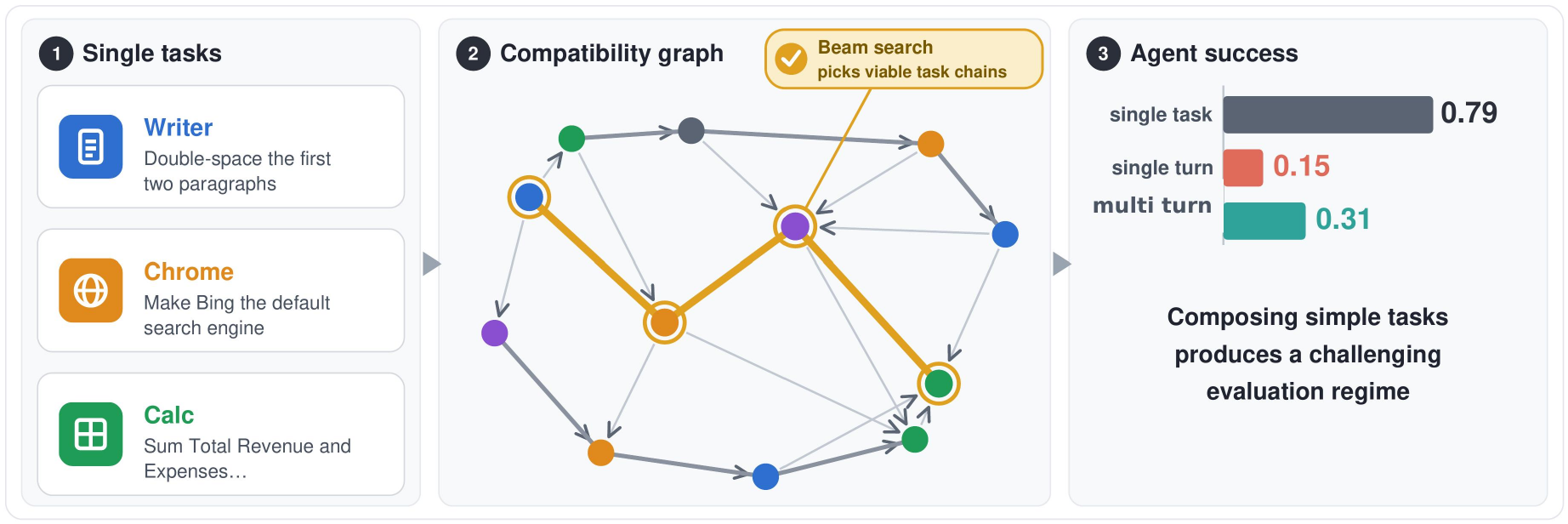}
\caption{Overview of ChainWorld workload construction and evaluation. The pipeline begins from atomic OSWorld tasks, builds a compatibility graph over task pairs, and selects a valid ordered chain. The resulting workload instance is evaluated in two protocols over the same underlying task sequence. Single turn evaluation presents the full chain in one prompt, while multi turn evaluation reveals one task at a time across turns while preserving state. In panel 3, the three bars correspond to the published GPT 5.5 atomic OSWorld reference \cite{singh2026openaigpt5card}, single turn ChainWorld completion, and multi turn ChainWorld completion. The atomic reference is shown for context and is not a subset matched comparison. Figure~\ref{fig:worked_example_chain} shows a concrete accepted chain rendered under both settings.}
\label{fig:chained_sessions_overview}
\end{figure*}

Computer use agents have improved on public benchmarks for web, mobile, and desktop interaction \cite{zhou2024webarena,koh2024visualwebarena,rawles2024androidworlddynamicbenchmarkingenvironment,OSWorld}. Most of these evaluations still measure one task at a time. That setup measures perception, interface control, and short horizon decision making, but only weakly measures longer horizon desktop behavior. In realistic desktop work, an agent may need to carry state across several related objectives and applications before the workload is complete. Doing so requires preserving state, tracking unfinished objectives, and moving across applications without losing context.

Atomic task performance therefore does not imply reliable longer horizon execution. This motivates evaluation settings that remain close to public desktop benchmarks while exposing multiobjective desktop behavior.

Two common alternatives are manual authoring and freeform synthesis. Manual authoring can produce high quality long horizon tasks, but it is expensive and difficult to scale with benchmark growth. Freeform language model synthesis scales more easily, but it shifts coherence, executability, and verifier validity into the generation pipeline itself.

Composition offers a different tradeoff. Public desktop benchmarks already contain atomic tasks with known setup, interface context, and evaluators. When state, snapshot, and evaluator constraints are satisfied, each task can participate in many compatible continuations. The existing task pool can therefore be expanded into longer horizon evaluation without reauthoring workflows from scratch. As new source tasks are added, the reachable workload space grows on top of the same benchmark infrastructure.

Recent work has studied generated trajectories, reverse task synthesis, and synthetic workloads as ways to expand supervision and probe capabilities that are difficult to collect directly \cite{sun2025osgenesisautomatingguiagent,xu2025agenttrekagenttrajectorysynthesis,murty2025nnetnavunsupervisedlearningbrowser,ou2024synatraturningindirectknowledge,pahuja2025explorerscalingexplorationdrivenweb,he2026efficientagenttrainingcomputer}. AgentSynth extends this direction to larger composed computer-use tasks at scale \cite{xie2026agentsynthscalabletaskgeneration}. Less developed is the evaluation side: how to characterize coordinated session behavior, where it breaks down, and how those failures vary with workload structure and task presentation \cite{zhuge2024agentasajudgeevaluateagentsagents,pan2024autonomousevaluationrefinementdigital,ma2024agentboard}.

We therefore compose existing OSWorld tasks under explicit state, snapshot, and evaluator constraints. Naive chaining can fail through snapshot mismatch, destructive state carryover, ambiguous artifact references, or evaluator assumptions that no longer hold. Our composition method is designed to produce longer workloads that remain executable, valid under the benchmark evaluators, and interpretable while remaining anchored to public task definitions and benchmark evaluators. In the final 320 case coded failure sample, runtime and setup failures account for only 3\% of failures, supporting the executability of the composed workload.

We represent tasks as nodes in a directional compatibility graph, search that graph under explicit structural targets, and allocate LLM judge calls to improve coverage across the candidate space. This preserves executable state continuity, screens out chains that do not read as one coherent desktop workload, and keeps coverage over workload structure explicit. Because each compatible task has many valid continuations, the same source pool can support larger workloads. In this paper, we instantiate that engine on OSWorld as a public held out evaluation setting. Figure~\ref{fig:chained_sessions_overview} summarizes the construction process and the two evaluation renderings used throughout the paper.

The resulting workload is built from public OSWorld tasks. ChainWorld joins compatible atomic tasks into desktop workloads of length two, three, or four while varying the underlying application footprint. This setup varies workload structure while keeping the task ingredients fixed, allowing us to measure how performance changes under composition, how performance changes between single turn and multi turn protocols, and how performance varies with workload structure \cite{laban2025llmslostmultiturnconversation}. Although each workload instance contains only a few atomic tasks, composition creates longer task specifications and stronger cross task state dependencies than the source tasks in isolation. The core methodological question is how to make workloads longer without degrading the quality of the benchmark itself.

Empirically, chain completion remains low. Multi turn evaluation improves completion for most models, but completion remains limited even when tasks are revealed one at a time. Aggregate scores also obscure variation across workload structure and protocol specific failure modes.

Our contributions are:
\begin{itemize}
    \item \textbf{ChainWorld composition and quality control.} We introduce ChainWorld, a constrained composition method that builds long horizon desktop workloads from atomic benchmark tasks while preserving the source evaluators, together with a three-stage quality control stack that separately enforces executable continuity, evaluator validity, and session coherence.
    \item \textbf{Paired protocol study.} We use paired single turn and multi turn renderings of the same chains to study how the gap between atomic task success and chained execution changes with task presentation.
    \item \textbf{Structured workload analysis.} We characterize performance under controlled variation in workload structure and failure mode, treating chained desktop evaluation as a structured workload rather than a single aggregate score.
\end{itemize}

\section{Related Work}

Computer use agent evaluation now spans web, mobile, and desktop environments \cite{zhou2024webarena,koh2024visualwebarena,rawles2024androidworlddynamicbenchmarkingenvironment,OSWorld}, but much of this work still evaluates one task at a time. We focus on long horizon desktop workloads composed from public atomic tasks.

\subsection{Benchmarks and Generated Workloads}

OSWorld is the main reference point because it provides a public desktop environment with visual interaction, file based state, and execution based evaluation \cite{OSWorld}. Spider2-V studies multi-application workflows inside OSWorld \cite{2024-spider2v}, while WorkArena++ and TheAgentCompany emphasize longer enterprise-style tasks \cite{boisvert2025workarenacompositionalplanningreasoningbased,xu2025theagentcompanybenchmarkingllmagents}. OdysseyBench and AgentSynth are closest in spirit to our setting \cite{wang2025odysseybenchevaluatingllmagents,xie2026agentsynthscalabletaskgeneration}. OdysseyBench synthesizes de novo workflows with broader narrative structure, while AgentSynth scales task generation through LLM-authored subtasks. AgentSynth generates new subtasks, while ChainWorld searches over existing evaluator backed tasks. We compose existing OSWorld tasks under directional compatibility and shared evaluator constraints, giving tighter control over evaluability and a fixed workload for paired protocol comparison.

Recent work also shows that task and trajectory generation choices can materially change downstream behavior \cite{sun2025osgenesisautomatingguiagent,xu2025agenttrekagenttrajectorysynthesis,murty2025nnetnavunsupervisedlearningbrowser,ou2024synatraturningindirectknowledge,pahuja2025explorerscalingexplorationdrivenweb,he2026efficientagenttrainingcomputer}. Here, generation serves a different purpose. We compose a long horizon evaluation workload from existing benchmark tasks rather than producing a training corpus.

\subsection{Verification and Long-Horizon Control}

Verification remains a persistent challenge for computer use evaluation. Rule based checks can be precise but may fail when agents reach correct outcomes through unanticipated paths, so recent work has explored more flexible judging and finer grained progress signals \cite{zhuge2024agentasajudgeevaluateagentsagents,pan2024autonomousevaluationrefinementdigital,ma2024agentboard}. In our setting, final success checks remain the source task evaluators, but intermediate states can vary across task order and protocol. The pipeline therefore uses plausibility filtering and downstream validation rather than naive concatenation alone.

Recent work has also focused on multi turn performance. Systems have improved through trajectory augmentation, process supervision, and more stable interaction \cite{he2026efficientagenttrainingcomputer,chen2025guishepherdreliableprocessreward,wang2025uitars2technicalreportadvancing,lai2025computerrlscalingendtoendonline}. Evidence also suggests that multi turn interaction can itself change model behavior even when the underlying problem is fixed \cite{laban2025llmslostmultiturnconversation}. Our single turn and multi turn settings probe that distinction within a common composed desktop workload.

\section{Methodology}

\subsection{Composition Engine}

We cast workload construction as constrained search over executable task compositions under a finite judging budget. The construction problem is to lengthen workloads while preserving executability, evaluator validity, and coherence.

Our method follows a decomposition of workload quality. A long horizon workload is useful for evaluation only if it is executable from a single starting state, valid under the composed evaluators, and coherent as one desktop session. Evaluator validity means that all source task evaluators can still be applied to the final chain state. We enforce these properties with separate mechanisms: compatibility rules for executable continuity, full chain validation for evaluator validity, and tuple level judging for session coherence. This factorization makes each quality condition auditable during construction.

Starting from atomic OSWorld tasks, the engine builds a directional compatibility graph over task pairs, searches that graph for structurally valid chains, and allocates judging effort to maximize coverage over workload structure while preserving evaluability. Here we instantiate the method on OSWorld to build ChainWorld, a held out long horizon desktop workload \cite{sun2025osgenesisautomatingguiagent,xu2025agenttrekagenttrajectorysynthesis,pahuja2025explorerscalingexplorationdrivenweb,xie2026agentsynthscalabletaskgeneration}.

\begin{figure*}[t]
\centering
\includegraphics[width=\textwidth]{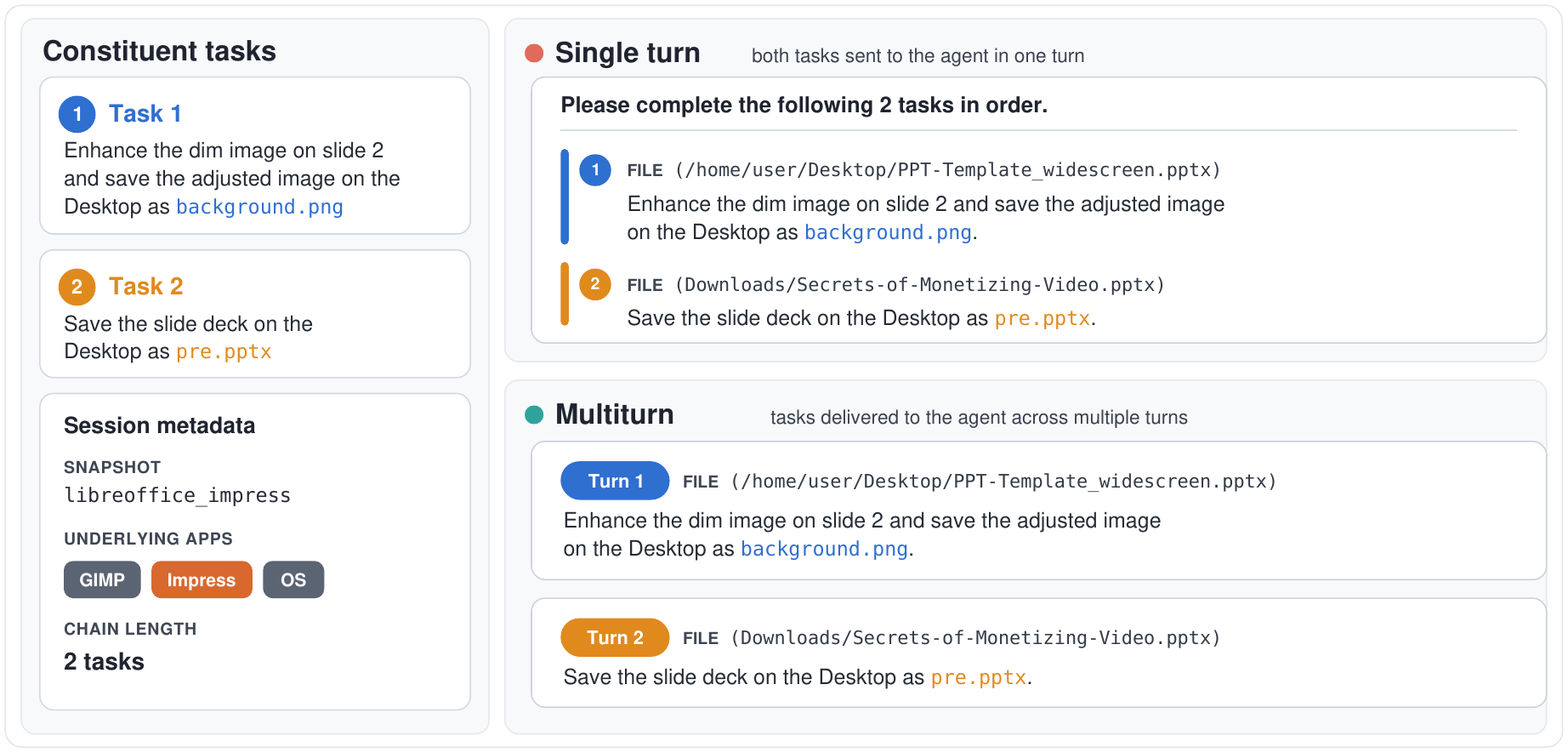}
\caption{Worked example of one ChainWorld workload. The figure shows the constituent tasks, the shared workload metadata, and the two agent facing renderings used in evaluation. In single turn evaluation, both tasks are presented together in one prompt. In multi turn evaluation, the same tasks are revealed across two turns while the desktop state and file references carry forward between turns.}
\label{fig:worked_example_chain}
\end{figure*}

\textbf{Task representation.}
The engine begins with deterministic per-task profiles extracted from the OSWorld task JSON. Each profile records the source application, related applications, snapshot, instruction text, setup actions, file and window writes, destructive operations, evaluator structure and reads, and whether the task is terminal-only. From the 369 OSWorld tasks, we exclude 8 Google Drive tasks from the composition pool to match the no-Drive evaluation setting used in the runs. We then annotate the remainder with additional fields for required applications, logged-in services, preexisting files, system state, end-state assertions, user goal, network dependencies, destructive intent, and category tags. These semantic fields are used only for workload construction rather than agent evaluation. They make executable state explicit, so the engine can reason about whether one atomic task depends on files, accounts, or application state that a previous task might destroy or overwrite.

\textbf{Pairwise compatibility graph.}
Candidate chains are assembled through a pairwise compatibility graph rather than by sampling workflows freely from a language model. A synthesized chain can appear plausible while still breaking the benchmark through snapshot mismatch, destroyed state, missing artifacts, evaluator interference, or file references that no longer resolve cleanly after earlier tasks.

Compatibility is directional. The score for $A \rightarrow B$ reflects whether task $B$ remains executable and evaluable after task $A$. The compatibility predicate first applies hard rules that reject pairs which cannot plausibly coexist in one session, including snapshot mismatch, terminal only placement, and unrecovered process kills. It then adds softer penalties and rewards that capture state reuse and likely interference, including cases where one task writes a file that another reads, browser state interference, same application continuity, and cases where one task produces a file later consumed by another.

The specific rule set is benchmark specific because it refers to OSWorld snapshots, setup actions, and evaluator structure, but the broader framework of directional compatibility scoring, constrained search, and tuple level filtering could transfer to benchmarks with analogous snapshot, setup, and evaluator metadata.

These constraints reduce infrastructure level confounds that would otherwise dominate the evaluation. Unconstrained chains often fail for setup reasons rather than agent behavior. Later tasks may require a different snapshot, an earlier task may kill a needed process or wipe required state, or prompts and evaluators may become ambiguous about which artifact a subtask refers to. Appendix Table~\ref{tab:snapshot_distribution} makes this constraint visible in the final workload.

\textbf{Constrained chain search.}
We search this graph with beam expansion over target workload cells defined by chain length and underlying application count. Search returns fixed ordered tuples that satisfy the requested cell by construction, so task order is determined before any judge call. Beam ranking uses cumulative compatibility score together with a light diversity bonus so that low frequency application families remain visible in the candidate pool.

\textbf{Workload ranges.}
We target chain lengths from 2 to 4 and underlying application counts from 1 to 3. Lengths 3 and 4 are long enough to require persistent session management, but still short enough that the resulting workloads remain recognizably composed from the source tasks rather than becoming open ended project work. The application count cells balance coverage between within application and cross application workloads during construction. Larger counts would force the engine toward rarer and often more contrived mixtures. The final 347-chain workload fills all nine target length by application-count cells, with 94--134 chains per length bucket and 100--127 chains per application-count bucket (Appendix Table~\ref{tab:chain_distribution}).

\textbf{Coverage-aware budgeted selection.}
The corpus builder treats judging effort as a limited budget and allocates it adaptively across candidate chains. We introduce coverage metrics to control how evenly the accepted set fills the target workload structure, including chain length, application count cells, source applications, and snapshots. A candidate gains coverage when it fills an underrepresented length, application count, source application, or snapshot cell. These counts are combined with online judge acceptance estimates to rank candidates under a finite judging budget.

Concretely, each application has a smoothed Beta Bernoulli acceptance estimate updated from prior judge outcomes, and each candidate is ranked by combining coverage gain, mean acceptance estimate across its applications, and compatibility score. In practice, this balances breadth against viable yield under a finite judging budget rather than sampling chains uniformly or at random.

\textbf{Tuple-level plausibility filtering.}
Each surviving candidate is then passed to a tuple level judge. The judge sees a compact representation of a fixed chain containing task identifiers, source applications, instructions, terminal only flags, and a brief summary of the task annotations when available. It returns a chain identifier, a one sentence narrative, brief risk notes, a binary identity consistency label, and a coarse naturalness score retained for profiling.

Executable continuity does not guarantee workload coherence. A chain can preserve state mechanically while still reading like a forced or incoherent desktop workload. Identity consistency asks whether the tasks could plausibly come from one consistent user session. It is the hard filter used to reject incompatible chains, while naturalness is recorded for profiling only. By the time a candidate reaches this stage, it has already passed compatibility screening, beam ranking, and coverage aware selection.

In the current pipeline, Opus 4.7 is used for both task annotation and tuple level judging. This role is limited to workload construction before any agent evaluation. The only source-pool exclusion is the 8 Google Drive tasks noted above. Opus does not reject source tasks. Instead, it operates at the chain level after search has proposed fixed tuples. All four models are then run on the same fixed evaluation set under the OSWorld-based evaluators. Because search fixes the candidate tuple before any judge call, the judge only accepts or rejects fixed tuples and annotates them \cite{zhuge2024agentasajudgeevaluateagentsagents,pan2024autonomousevaluationrefinementdigital}.

In the final build, 9 of 389 judged candidate chains are rejected for identity inconsistency. This 97.7\% pass rate indicates that the compatibility graph and coverage-aware selection already do most of the coherence work before the judge is applied. Together with the compatibility predicate and full chain validation, this creates a three-stage quality control stack with local executability, chain level plausibility, and full chain state validation. Appendix Table~\ref{tab:pipeline_stats} reports the construction totals, Appendix Table~\ref{tab:chain_distribution} gives the exact length-by-application breakdown of the accepted workload, and Appendix Table~\ref{tab:snapshot_distribution} reports the snapshot distribution.

\textbf{Validation and export.}
Accepted tuples pass through full chain validation and set based deduplication before they enter the final evaluation set. Validation checks cumulative state constraints over the entire chain rather than only adjacent pairs. When a longer candidate fails validation, the engine trims the tail and retries the shorter prefix, allowing a failed length 4 candidate to yield a valid length 3 or length 2 chain instead of being discarded outright. The final evaluation set contains 347 accepted chains spanning lengths 2 to 4 and underlying application counts 1 to 3. Across those chains, 189 of the 361 OS World tasks appear in at least one accepted chain.

Coverage is therefore broad but selective rather than exhaustive. Among tasks that never appear in an accepted chain, terminal only tasks and destructive or state modifying tasks are modestly overrepresented. Eighteen of 180 unused tasks are terminal only, compared with 11 of 189 used tasks, and 16 of 180 unused tasks contain destructive intent or destructive operations, compared with 10 of 189 used tasks. This pattern is consistent with the design of the compatibility predicate. Tasks with few continuations that preserve state remain in the source pool, but they enter accepted chains less often.

\subsection{Evaluation Protocol}

\textbf{Single turn and multi turn settings.}
We evaluate two protocols over the same accepted chains. In the single turn setting, the chain is converted into a single OSWorld-shaped task whose setup is the concatenation of all constituent setup actions and whose evaluator is the conjunction of all constituent evaluators. The single turn instruction is not rewritten by a language model. Instead, each underlying task instruction is augmented deterministically with file path context derived from the task profile, and the resulting turn texts are concatenated into an ordered prompt. We do this because some source tasks refer indirectly to their artifacts, and paraphrasing could introduce avoidable ambiguity about which artifact belongs to which subtask.

Single turn evaluation therefore uses one combined prompt, one initial snapshot, and a total budget of 75$n$ steps for an $n$-task chain. In the multi turn setting, the same per task instructions are revealed one at a time. Later turns apply their setup on top of the current desktop state without another snapshot revert while rebinding the evaluator to the next task. Each turn is evaluated immediately after that turn, and chain completion requires every per turn evaluator to succeed. Later turns can change the desktop state after earlier scores have been recorded. Multi turn evaluation uses multiple user turns, persistent state across turns, and 75 steps for each revealed task.

The comparison keeps the underlying task sequence and total interaction budget fixed. It isolates two differences. One is simultaneous versus sequential revelation of the same task sequence. The other is flexible versus per task capped budget allocation. Figure~\ref{fig:worked_example_chain} shows a compact example of one accepted chain and its renderings.

\textbf{Run configuration.}
All runs use the same Ubuntu desktop environment in Docker, the same \texttt{pyautogui} action space, screenshot observations, and a 1920 by 1080 screen. The protocol level comparison aligns the total budget at 75 steps per constituent task in both settings. Single turn chains receive a length scaled session budget that can be spent anywhere in the chain, while multi turn chains receive 75 steps for each revealed task. Reported scores come from a single run for each model, chain, and protocol combination. Model specific decoding and API settings follow the recorded run configurations rather than the workload design itself. Kimi 2.5 uses image history capped at one screenshot due to API constraints.

\textbf{Metrics.}
We report two metrics. Chain completion marks a run correct only when every task in the chain succeeds. Task success measures the fraction of constituent tasks completed across all chains. Chain completion is the main outcome because end to end workload completion is the capability we want to measure. Task success is a companion outcome that separates total failure from partial progress. In the single turn setting, the compound evaluator is built by concatenating the original verifier lists for all tasks, and per task success is recovered by partitioning the verifier outputs back into the original task groups. In the multi turn setting, per task success is read from the evaluator score recorded after each turn. This distinction allows us to compare chain completion against local task progress under a common workload design \cite{ma2024agentboard}. We evaluate GPT 5.5, Opus 4.7, Kimi 2.5, and Gemini 3.1 Pro under this common protocol \cite{singh2026openaigpt5card,anthropic2026opus,kimiteam2026kimik25visualagentic,googledeepmind2026gemini31pro}.

\section{Results}
 
\subsection{Overall Performance}
\begin{figure*}[!t]
\centering
\includegraphics[width=\textwidth]{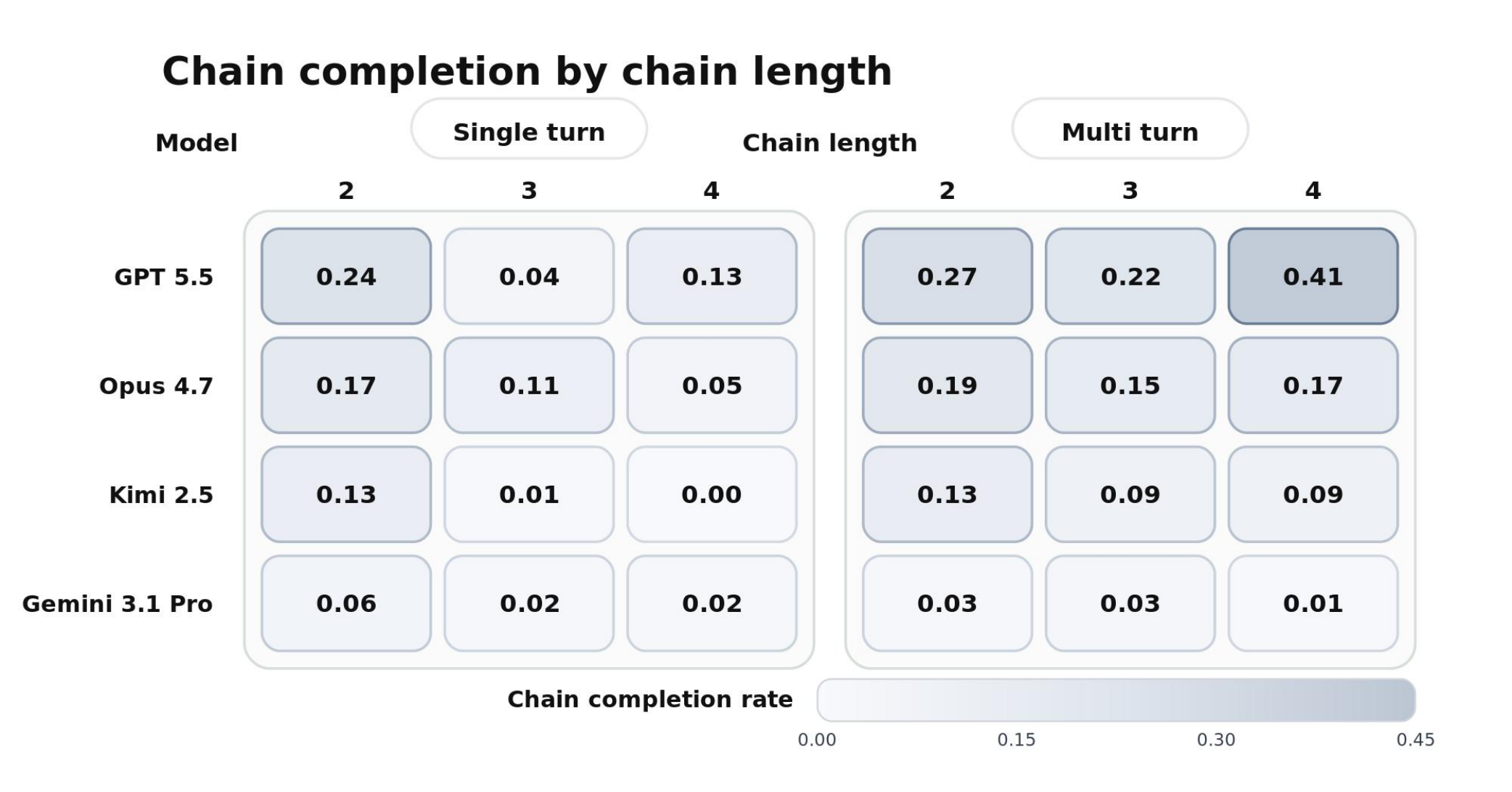}
\caption{Chain completion by chain length. Each cell reports the fraction of chains completed when the workload contains two, three, or four constituent tasks. The left block shows single turn evaluation, where the full chain is presented at once; the right block shows multi turn evaluation, where the same underlying chains are revealed one task at a time across turns. Multi turn improves completion for GPT 5.5, Opus 4.7, and Kimi 2.5, while Gemini 3.1 Pro remains low in both settings. Section~4.1 reports bootstrap 95\% confidence intervals for the overall completion rates.}
\label{fig:length_heatmap}
\end{figure*}

Table~\ref{tab:overall_results} summarizes the main results. Across all four models, maximum chain completion is 31\% on the ChainWorld workload. Bootstrap intervals in the table are computed by resampling chains within each model-protocol condition, so they reflect chain variation within a single run rather than run to run variation. GPT 5.5 is strongest in both settings, Opus 4.7 follows, and Kimi 2.5 and Gemini 3.1 Pro are weaker.

\begin{table}[!b]
\centering
\footnotesize
\setlength{\tabcolsep}{2.5pt}
\begin{tabular}{lcccc}
\toprule
 & \multicolumn{2}{c}{Chain completion} & \multicolumn{2}{c}{Task success} \\
\cmidrule(lr){2-3}\cmidrule(lr){4-5}
Model & Single & Multi & Single & Multi \\
\midrule
GPT 5.5 & 0.15{\scriptsize $\pm$0.04} & 0.31{\scriptsize $\pm$0.05} & 0.49 & 0.63 \\
Opus 4.7 & 0.11{\scriptsize $\pm$0.03} & 0.17{\scriptsize $\pm$0.04} & 0.32 & 0.50 \\
Kimi 2.5 & 0.05{\scriptsize $\pm$0.02} & 0.10{\scriptsize $\pm$0.03} & 0.23 & 0.41 \\
Gemini 3.1 Pro & 0.03{\scriptsize $\pm$0.02} & 0.02{\scriptsize $\pm$0.02} & 0.18 & 0.23 \\
\bottomrule
\end{tabular}
\caption{Overall ChainWorld results by model and protocol. Chain completion requires all tasks in the chain to succeed. Uncertainty terms report bootstrap 95\% confidence intervals from chain resampling. Task success is the fraction of constituent tasks completed across all chains.}
\label{tab:overall_results}
\end{table}

Multi turn chain completion is higher than single turn completion for GPT 5.5, Opus 4.7, and Kimi 2.5. On shared chains, exact McNemar tests support the gain for all three models with $p < 0.004$, but do not show a multi turn gain for Gemini 3.1 Pro. The two settings run on the same underlying chains with the same total interaction budget for a given chain length. This protocol contrast reflects both task revelation and budget allocation flexibility. For context, panel 3 of Figure~\ref{fig:chained_sessions_overview} includes the published GPT 5.5 atomic OSWorld reference \cite{singh2026openaigpt5card}. The bar provides an atomic benchmark reference and is not a subset matched comparison.

Task success remains higher than chain completion, which indicates that failed chains often contain local task progress rather than immediate collapse. Chained evaluation therefore separates local task completion from end to end workload completion, a distinction that atomic task metrics can obscure \cite{ma2024agentboard}.

\subsection{Variation Across Chain Length}

Figure~\ref{fig:length_heatmap} shows that chain length is informative, but not a simple monotonic difficulty axis. The clearest decline appears in the single turn setting, where Opus 4.7 drops from 0.17 at length 2 to 0.05 at length 4 and Kimi 2.5 drops from 0.13 to 0.00. Gemini 3.1 Pro stays near the floor across lengths. GPT 5.5 shows a different pattern, dropping from 0.24 at length 2 to 0.04 at length 3 and then partially recovering to 0.13 at length 4. This may reflect workload mix, but given the modest per-length counts and single-run design, we treat it as suggestive rather than stable evidence of a length-specific effect. We therefore do not treat length as a standalone difficulty variable.

Multi turn evaluation improves performance for most models, but it does not remove variation across chain length. Sequential task revelation helps, yet the ChainWorld workload still separates models by their ability to sustain a multi-objective workload across several coordinated tasks.

\subsection{Failure Modes}

To better understand how chains fail, we analyze a stratified coded sample of 320 failed runs drawn uniformly across the four models and two protocols (40 per model-protocol cell). Each case is assigned a single primary failure mechanism via an LLM judge rubric, with ambiguous cases rechecked under a stricter prompt. Appendix~\ref{app:failure_modes} gives the full setup, and Appendix Table~\ref{tab:failure_modes_breakdown} gives the model-by-protocol breakdown.

\begin{table*}[t]
\centering
\normalsize
\setlength{\tabcolsep}{5pt}
\begin{tabular}{p{0.3\textwidth} p{0.4\textwidth} r r r}
\toprule
Category & Example & Overall & Single & Multi \\
\midrule
Near-miss artifact error & File created, wrong value or format. & 60\% & 72\% & 47\% \\
Early-subtask fixation & Keeps working on Task 1, and later tasks barely begin. & 10\% & 8\% & 12\% \\
Fragmented progress & Several tasks are attempted, but all remain partially complete. & 15\% & 7\% & 22\% \\
Later-turn disengagement & Later turns get 0 or 1 actions before agent declares DONE. & 8\% & 1\% & 15\% \\
Tool/debugging drift & Writes scripts or probes instead of outputs. & 3\% & 6\% & 1\% \\
Runtime/setup failure & File-open or API failure stops the run. & 3\% & 4\% & 2\% \\
Other & Residual idiosyncratic failures. & 1\% & 2\% & 0\% \\
\bottomrule
\end{tabular}
\caption{Primary failure-mode labels for the coded failure sample. Percentages are computed over all coded failures and separately within single turn and multi turn failures. Single turn failures most often involve near-miss artifact errors, while multi turn failures are more heterogeneous.}
\label{tab:failure_modes}
\end{table*}

The most common pattern overall is near miss execution. Most failures are not empty trajectories or complete task abandonment. Instead, the agent often produces the relevant file or output but misses exact content, formatting, or verifier critical details.

The failure profile also differs by protocol. Single turn evaluation most often involves near miss artifact errors, while multi turn evaluation is more heterogeneous. Fragmented progress accounts for 22\% of multi turn failures and 7\% of single turn failures, and later turn disengagement rises from 1\% in single turn to 15\%. Fragmented progress spreads effort across several tasks without completing them, whereas disengagement reflects a sharp drop in effort once later turns arrive. Tool and debugging drift shows the opposite pattern, accounting for 6\% of single turn failures but only 1\% of multi turn failures. Single turn evaluation more often fails at producing an exact end to end artifact bundle, while multi turn evaluation shows a broader range of ways in which a session can stall or drift.

Runtime and setup failures are rare in both settings, which supports two readings. At the model level, many failures reflect agent behavior rather than infrastructure breakdown. At the workload level, the composition pipeline produces executable workloads. If the composed tasks were dominated by mechanically incompatible chains, setup and infrastructure failures would be much more common.

The two protocols are therefore informative in different ways. Single turn evaluation stresses final state precision under a single combined specification. In near miss cases, the agent often reaches the right application state and produces the right kind of artifacts while missing exact verifier critical details. Multi turn evaluation stresses workload management more directly.

\subsection{Implications for Evaluation}

Taken together, these results distinguish local task progress from full chain completion. Agents often perform useful local work inside a chain, yet fail to complete the whole workload. Atomic task evaluation can miss this distinction because success is measured one objective at a time.

Aggregate scores also obscure structure. Chain completion varies by chain length and by evaluation protocol, so a single number does not summarize the workload well. Session length, task composition, and prompt rendering all influence what kind of failure is being measured. Near miss failures suggest that agents often produce relevant outputs but miss verifier critical details, which has different implications for evaluation and training than coordination failures such as fixation or disengagement.

The failure taxonomy also suggests different forward paths for training and system design. Near miss artifact errors point toward stronger supervision on artifact correctness and reward signals that distinguish close but wrong outputs from no useful progress. By contrast, fixation, disengagement, and fragmented progress point toward workload management mechanisms such as planning, progress summaries, and stronger support for carrying unfinished objectives across a multi-objective workload.

These findings also matter for future training work. As task and trajectory synthesis pipelines become more scalable, ChainWorld can help identify which coordination failures persist under composed workloads. The variation across chain length and protocol suggests that optimizing only for isolated outcomes may not be sufficient to sustain coordinated behavior across longer workloads.

\section{Conclusion}

ChainWorld composes atomic OSWorld tasks into executable long horizon workloads under explicit compatibility and evaluator constraints. Across four current computer use agents, it exposes a substantial gap between local task progress and coordinated desktop execution, with single turn and multi turn protocols revealing different behavior. The main implication is that long horizon desktop evaluation can be built from evaluator backed atomic tasks while revealing coordination failures that atomic task scores do not capture.

\section{Limitations}

This study is scoped to composed desktop workloads built from chains of two to four OSWorld tasks. These workloads are longer-horizon than the atomic source tasks, but they remain more structured than open-ended desktop assistance, and the reported results should be interpreted within that regime. The composition method is also benchmark-specific in one important respect: the compatibility predicate is instantiated from OSWorld snapshots, setup actions, and evaluator structure. The broader framework could transfer to benchmarks with analogous metadata, but we do not demonstrate that transfer here.

The experiments use one run per model, chain, and protocol, so the paper reports structured coverage over the workload rather than per-chain variance estimates. This limits claims about run-to-run variability.

Opus 4.7 is also used during workload construction for task annotation and tuple-level judging, and is then evaluated as one of the target models. The evaluation workload is fixed before any model runs and is shared across all four models. GPT 5.5 is still strongest under both protocols, so this overlap does not trivially favor Opus 4.7 in the final workload. The direct chain-level filtering role is narrow, with 9 identity-consistency rejections out of 389 judged candidate chains, but the overlap can still introduce construction bias that future robustness checks should test more directly. The naturalness ratings in Appendix~\ref{app:coherence_profile} are produced by the same judge model used during construction, so they should be read as internal consistency rather than independent quality verification. Independent human inter-rater agreement remains a natural next step. 

Potential risks are primarily interpretive. Failures on this workload should not be read as a full account of deployment risk for computer-use agents, and the converse is also true: strong performance on these composed workloads would not establish safety or robustness in open environments. The workload is intended as a controlled evaluation slice for longer-horizon desktop execution, not as a complete model of real-world desktop use.

\bibliographystyle{abbrvnat}
\bibliography{custom}

\clearpage
\appendix
\section{Workload Construction Details}
\label{app:construction_details}

\subsection{Pipeline Statistics}
\label{app:construction_stats}

An anonymous release of the construction code and workload metadata is available at \url{https://anonymous.4open.science/r/os_world_chained_review-E8D6}. The release includes the composition engine, accepted workload metadata, and judging schema used for the public evaluation set.

Table~\ref{tab:pipeline_stats} summarizes the logged workload-construction totals across the initial build and subsequent append passes that produced the final 347-chain evaluation set. These counts show selective construction rather than enumeration and retention of all graph candidates. Most pooled candidates are never judged, and the tuple level judge is applied to a relatively small screened subset. The high acceptance rate at that stage reflects upstream screening rather than judge redundancy. By the time a candidate reaches the judge, compatibility filtering, beam ranking, and coverage-aware selection have already removed most weak tuples. Identity consistency rejects only 9 of 389 judged candidates, which is consistent with the main workload shaping already happening upstream.

\begin{table}[h]
\centering
\small
\resizebox{0.6\columnwidth}{!}{%
\begin{tabular}{lrr}
\toprule
Statistic & Count & Share \\
\midrule
Candidate tuples pooled & 5,296 & -- \\
Pre-judge dedup rejections & 589 & 11.1\% of pooled \\
Tuple-level judge calls & 389 & 7.3\% of pooled \\
Identity-consistency rejections & 9 & 2.3\% of judged \\
Accepted chains & 347 & 89.2\% of judged \\
Trim recoveries & 148 & 42.7\% of accepted \\
\bottomrule
\end{tabular}
}
\caption{Aggregated workload-construction statistics from the logged build and append passes that produced the final evaluation workload. ``Candidate tuples pooled'' counts graph search candidates considered by the builder, ``tuple level judge calls'' counts candidates actually sent to the judge, and ``trim recoveries'' counts longer candidates that failed full-chain validation but yielded a shorter valid prefix.}
\label{tab:pipeline_stats}
\end{table}

\begin{table}[t]
\centering
\small
\begin{tabular}{lrrrr}
\toprule
Chain length & 1 app & 2 apps & 3 apps & Total \\
\midrule
2 & 41 & 38 & 40 & 119 \\
3 & 17 & 40 & 37 & 94 \\
4 & 42 & 42 & 50 & 134 \\
\midrule
Total & 100 & 120 & 127 & 347 \\
\bottomrule
\end{tabular}
\caption{Distribution of accepted chains in the final evaluation set by chain length and underlying number of applications. Application counts are reconstructed from task profiles, so chains whose coarse labels pass through aggregate buckets such as \texttt{multi\_apps} still resolve to their underlying app sets.}
\label{tab:chain_distribution}
\end{table}

\begin{table}[t]
\centering
\small
\resizebox{0.5\columnwidth}{!}{%
\begin{tabular}{lrr}
\toprule
Snapshot & Source tasks & Accepted chains \\
\midrule
os & 31 & 67 \\
libreoffice\_calc & 74 & 56 \\
thunderbird & 20 & 40 \\
libreoffice\_impress & 51 & 31 \\
multiapps & 13 & 28 \\
chrome & 72 & 27 \\
libreoffice\_writer & 31 & 25 \\
gimp & 32 & 23 \\
base\_setup & 11 & 19 \\
vlc & 4 & 18 \\
vscode & 30 & 13 \\
\bottomrule
\end{tabular}
}
\caption{Snapshot distribution of the source task pool and accepted ChainWorld workloads. Source-task counts are computed after exclusions. Accepted-chain counts use the final chain snapshot field. Snapshot aliases are normalized by merging \texttt{vs\_code} into \texttt{vscode} and \texttt{multi\_apps} into \texttt{multiapps}.}
\label{tab:snapshot_distribution}
\end{table}

\subsection{Snapshot Profile}
\label{app:snapshot_profile}

Table~\ref{tab:snapshot_distribution} helps clarify what the snapshot constraint means in practice. Several snapshots are effectively application-specific. For example, the source pool under \texttt{libreoffice\_impress}, \texttt{libreoffice\_writer}, \texttt{libreoffice\_calc}, \texttt{thunderbird}, \texttt{gimp}, and \texttt{vscode} is dominated by tasks from the corresponding application family, even though some \texttt{multi\_apps} tasks also share those starting states. By contrast, \texttt{os}, \texttt{multiapps}, and \texttt{base\_setup} function as broader snapshots that support more cross-application continuations.

These counts show that snapshot compatibility is not a minor bookkeeping condition. It materially shapes which continuations are available. Many accepted workloads cluster within a small number of snapshot families, and even application-specific snapshots can still support some two-application chains when \texttt{multi\_apps} tasks share the same starting state. For this reason, snapshot matching appears as a hard compatibility condition in the graph rather than as a soft preference.

\subsection{Tuple-Level Judge Interface}
\label{app:judge_interface}

The tuple level judge operates on a fixed ordered candidate tuple after beam search. It does not compose, rewrite, or reorder tasks. For each task in the tuple, the judge receives a compact JSON object containing the task identifier, source application, original instruction text, terminal-only flag, and when available a short semantic summary derived during profile annotation (user goal, tags, and destructive intent). The judge returns five fields used by the pipeline: a short chain identifier, a one-sentence narrative, brief risk notes, a binary identity-consistency label, and a naturalness score.

The core of the system prompt is shown below.

\begin{quote}
\small
\textit{True iff a single real person could plausibly be the one requesting all of these tasks. People multitask freely, so loose context switching is fine; the bar is only that nothing contradicts an implicit consistent user.}

\textit{How naturally would the chain read as one user's session? This is independent of identity consistency. Score honestly across the full range from clearly coherent workflows to forced or unusual combinations.}
\end{quote}

This interface is deliberately narrow and conservative. The judge evaluates only the plausibility of a fixed tuple and records lightweight metadata about it. The search procedure, compatibility predicate, and validation logic determine which candidates are screened in and how they are ordered.

\subsection{Coherence Profile}
\label{app:coherence_profile}

The accepted workload targets coherent multi-objective desktop workloads rather than full real-world desktop coverage. The tuple level judge records a coarse naturalness score from 1 to 5 for each accepted chain. In the final 347-chain evaluation set, no accepted chains receive a score of 1, and 304/347 (88\%) receive scores in the 3--5 range. The distribution is 43 chains at score 2, 154 at score 3, 117 at score 4, and 33 at score 5. We treat these ratings as descriptive support rather than as the primary acceptance criterion. Together with the staged filtering statistics above and the main-text failure-mode analysis, where runtime and setup failures are rare, they support the intended construction objective. The accepted workloads are filtered toward chains that remain executable, valid under the benchmark evaluators, and coherent enough to support model evaluation.

As a sanity check on this rubric, we also scored an internal reference set of 370 human written long horizon desktop prompts with the same judge and the same 1--5 naturalness scale. This set is not part of ChainWorld and is not a matched benchmark comparison. It is used only as a manually authored reference point. In that set, all 370 prompts were identity consistent, the mean naturalness score was 4.51, the median was 5, and the score distribution was 2 at score 2, 6 at score 3, 162 at score 4, and 200 at score 5. These scores are higher than the composed ChainWorld distribution, which is expected for manually authored prompts, but they place the ChainWorld ratings in context. The composed workload sits below manual authoring while remaining concentrated in the 3--5 range with no score 1 chains.

\subsection{Pairwise Compatibility Scoring}
\label{app:compatibility_scoring}

The compatibility graph is built over ordered task pairs. Each pair is scored by a fixed rule set rather than by freeform workflow synthesis. The scoring has three layers. First, hard exclusions remove pairs that cannot be executed or evaluated coherently from one starting state, such as snapshot mismatch, terminal-only placement, or unrecovered process kills. Second, soft penalties flag likely interference, such as one task overwriting files or browser state that the later task or evaluator depends on. Third, positive rules reward clean continuity, such as same-application workflows, shared application footprint, or cases where one task produces a file later consumed by another. When semantic task annotations are available, the scorer also uses them to recognize downstream state reuse and destructive conflicts that are not visible from static parsing alone.

This scoring serves a different purpose from the tuple level judge. The graph score is a local structural filter and ranking signal. It decides whether one task can plausibly follow another while preserving evaluability. The tuple level judge operates only after beam search has already produced a fixed candidate chain.

\subsection{Failure-Mode Coding}
\label{app:failure_modes}

For the qualitative failure-mode analysis in the main text, we code a retained sample of 320 failed runs drawn across models and protocols. Each run is labeled with one primary category from a fixed taxonomy: near-miss artifact error, early-subtask fixation, fragmented progress, later-turn disengagement, tool or debugging drift, runtime or setup failure, and a small residual other category. The first pass uses GPT 5.4 mini as the judge. The coding prompt receives the chain-level result summary together with short trace and runtime-log excerpts. To avoid collapsing many qualitatively different cases into a generic progress label, 96 ambiguous cases are rejudged with GPT 5.4 under a stricter prompt that requires a concrete dominant mechanism.

\begin{table*}[t]
\centering
\small
\setlength{\tabcolsep}{4pt}
\begin{tabular}{l r r r r r r r}
\toprule
Model / protocol & Near-miss & Fixation & Fragmented & Disengage & Drift & Runtime & Other \\
\midrule
GPT 5.5 single turn & 88\% & 0\% & 0\% & 5\% & 8\% & 0\% & 0\% \\
GPT 5.5 multi turn & 72\% & 0\% & 18\% & 2\% & 0\% & 8\% & 0\% \\
Kimi 2.5 single turn & 62\% & 18\% & 0\% & 0\% & 8\% & 12\% & 0\% \\
Kimi 2.5 multi turn & 60\% & 10\% & 25\% & 5\% & 0\% & 0\% & 0\% \\
Opus 4.7 single turn & 78\% & 10\% & 0\% & 0\% & 8\% & 2\% & 2\% \\
Opus 4.7 multi turn & 35\% & 25\% & 15\% & 22\% & 2\% & 0\% & 0\% \\
Gemini 3.1 Pro single turn & 62\% & 2\% & 28\% & 0\% & 0\% & 0\% & 8\% \\
Gemini 3.1 Pro multi turn & 20\% & 15\% & 32\% & 30\% & 0\% & 2\% & 0\% \\
\bottomrule
\end{tabular}
\caption{Failure-mode breakdown by model and protocol. Percentages are computed within each retained model-protocol sample of 40 failures. ``Near-miss'' denotes runs that produce relevant artifacts but miss exact verifier-critical details, while ``Fragmented'' denotes runs that make progress on several tasks without bringing any one of them to completion. The table shows that single turn failures most often involve near-miss artifact errors, whereas multi turn failures, especially for Opus 4.7 and Gemini 3.1 Pro, are more mixed.}
\label{tab:failure_modes_breakdown}
\end{table*}

\end{document}